\documentclass[sigconf]{acmart}

\setcopyright{acmlicensed}
\copyrightyear{2018}
\acmYear{2018}
\acmDOI{XXXXXXX.XXXXXXX}
\acmConference[Conference acronym 'XX]{Make sure to enter the correct
  conference title from your rights confirmation email}{June 03--05,
  2018}{Woodstock, NY}

\usepackage{multirow}
\usepackage[utf8]{inputenc}
\usepackage{lipsum}
\usepackage{xcolor}
\usepackage{colortbl}
\usepackage{enumitem}
\usepackage{booktabs}
\usepackage{multirow}
\usepackage{graphicx}
\usepackage{xcolor}
\definecolor{positive}{RGB}{200, 35, 35}
\definecolor{negative}{RGB}{0, 130, 70}

\definecolor{paperblue}{rgb}{0.1,0.3,0.6}

\newcommand{\ourname}{A/B Agent}
\usepackage{booktabs}
\usepackage{multirow}
\usepackage{graphicx}
\usepackage[table]{xcolor}

\definecolor{lightgray}{RGB}{244,244,244}
\definecolor{rowgray}{RGB}{242,242,242}
\definecolor{mygray}{RGB}{232,232,232}
\definecolor{darkgray}{RGB}{218,218,218}

\definecolor{bestred}{RGB}{238,180,176}
\definecolor{secondred}{RGB}{248,211,208}
\definecolor{positivered}{RGB}{250,226,224}
\definecolor{negativegreen}{RGB}{226,240,228}

\begin{document}

\title{\ourname: A Self-Evolving Agent for Strategy Iteration in Industrial A/B Testing}

\author{Zhuohang Jiang}
\authornote{Work done during an internship at Kuaishou Technology.}
\affiliation{%
  \institution{The Hong Kong Polytechnic University}
  \city{Hong Kong SAR}
  \country{China}
}
\email{zhuohang.jiang@connect.polyu.hk}

\author{Yuxin Chen}
\authornote{Corresponding Author.}
\affiliation{%
  \institution{Kuaishou Technology}
  \city{Beijing}
  \country{China}
}
\email{chenyuxin06@kuaishou.com}

\author{Yongsen Pan}
\authornotemark[1]
\affiliation{%
  \institution{University of Electronic Science and Technology of China}
  \city{Chengdu, SiChuan}
  \country{China}
}
\email{panys@std.uestc.edu.cn}

\author{Boyang Xia}
\affiliation{%
  \institution{Kuaishou Technology}
  \city{Beijing}
  \country{China}
}
\email{xiaboyang.tech@gmail.com}

\author{Zheng Hu}
\authornotemark[1]
\affiliation{%
  \institution{Southwest Jiaotong University}
  \city{Chengdu, SiChuan}
  \country{China}
}
\email{huzheng@swjtu.edu.cn}

\author{Wenqi Fan}
\authornotemark[2]
\affiliation{%
  \institution{The Hong Kong Polytechnic University}
  \city{Hong Kong SAR}
  \country{China}
}
\email{wenqifan03@gmail.com}

\author{Qing Li}
\authornotemark[2]
\affiliation{%
  \institution{The Hong Kong Polytechnic University}
  \city{Hong Kong SAR}
  \country{China}
}
\email{csqli@comp.polyu.edu.hk}

\author{Hongyang Wang}
\affiliation{%
  \institution{Kuaishou Technology}
  \city{Beijing}
  \country{China}
}
\email{liboyu08@kuaishou.com}

\author{Jun Wang}
\affiliation{%
  \institution{Kuaishou Technology}
  \city{Beijing}
  \country{China}
}
\email{wangjun03@kuaishou.com}

\author{Zhongbo Sun}
\affiliation{%
  \institution{Kuaishou Technology}
  \city{Beijing}
  \country{China}
}
\email{szb17@tsinghua.org.cn}

\author{Wenwu Ou}
\affiliation{%
  \institution{Kuaishou Technology}
  \city{Beijing}
  \country{China}
}
\email{luocheng10@kuaishou.com}

\author{Peng Jiang}
\affiliation{%
  \institution{Kuaishou Technology}
  \city{Beijing}
  \country{China}
}
\email{jiangpeng11@kuaishou.com}

\renewcommand{\shortauthors}{Zhuohang Jiang et al.}

\begin{abstract}

Industrial recommendation strategy iteration heavily relies on large-scale A/B experimentation. Traditional tuning requires experts to repeatedly design strategies, configure experiments, analyze results, and adjust parameters, making the process labor-intensive and time-consuming. Meanwhile, valuable knowledge from historical experiments is often fragmented, making systematic reuse difficult through manual expert effort alone. Existing RAG agents partially alleviate this burden by retrieving prior strategies, but typically organize experience in a flat manner, overlooking the hierarchical relationships among business scenarios, recommendation stages, optimization objectives, and experimental contexts. This often results in mismatched retrieval and limited cross-scenario transfer, while preventing agents from continuously refining strategies and parameters through sequential A/B feedback.
To address these limitations, we propose \textbf{\ourname}, a closed-loop A/B agent for industrial recommendation strategy optimization. The framework comprises three tightly coupled core components: \textit{Historical Strategy Knowledge Organization}, \textit{Autonomous Target-Aware Strategy Generation}, and \textit{Experiment-Guided Strategy Self-Evolution}. It organizes historical strategies into a hierarchical experience tree, retrieves transferable evidence through multi-path Tree-RAG to generate executable strategies, and continuously analyzes online A/B feedback to guide autonomous tuning and update the experience tree for self-evolution. Extensive offline and online evaluations demonstrate its effectiveness, including a \textbf{4.829\%} improvement in GMV in a real-world short-video e-commerce recommendation system while maintaining positive gains across all guardrail metrics.

\end{abstract}



\maketitle
\section{Introduction}

In modern industrial recommender systems~\cite{pan2026beyond,hu2026stop,jiang2026atomic}, continuous strategy development~\cite{peng2025commerce}, large-scale A/B testing~\cite{quin2024b}, and feedback-driven parameter optimization~\cite{han2019optimizing} are essential for improving recommendation quality and performance. A strategy iteration~\cite{jadon2024comprehensive, deng2025onerec} typically spans objectives, ranking pipelines, scoring functions, parameter configurations, and engineering constraints, followed by small-scale validation, traffic ramp-up, and full deployment. Because multiple business domains share limited exposure, optimizing one objective may suppress others and harm user experience. Therefore, strategy iteration~\cite{gupta2019top} must jointly optimize core metrics and satisfy multiple guardrail constraints~\cite{pi2020search} to ensure balanced traffic allocation and long-term system health.

\begin{figure}
\centering
\includegraphics[width=\linewidth]{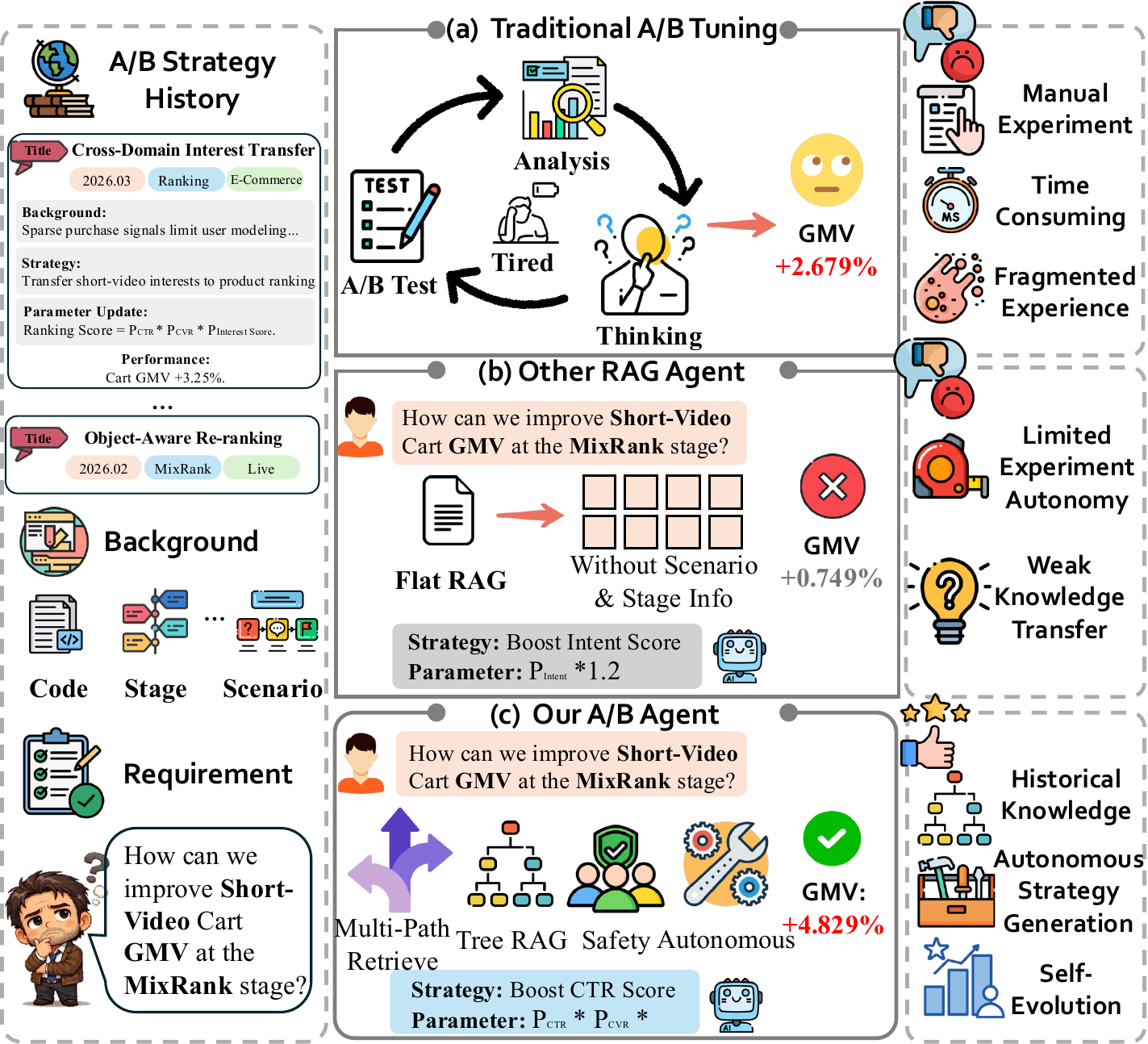}
\caption{Motivation of \ourname{}. Traditional A/B tuning requires repeated manual experimentation, while Other RAG Agent often misses scenario- and stage-specific context. In contrast, \ourname{} leverages hierarchical historical knowledge, autonomous strategy generation, and self-evolution to optimize strategies effectively.}
\label{fig:intro}
\end{figure}

However, industrial strategy optimization faces substantial challenges due to the high costs and long iteration cycles of online A/B testing, which result in sparse and delayed feedback for parameter optimization. Consequently, conventional reinforcement learning (RL)-based optimization methods~\cite{feng2025experimenting, letham2019constrained}, which typically require extensive environment interactions and repeated trial-and-error, are difficult to apply efficiently in industrial settings. As shown in Figure~\ref{fig:intro} (a), industrial strategy iteration therefore remains heavily dependent on human expertise~\cite{dmitriev2017dirty}. Faced with numerous interdependent and sometimes conflicting core and guardrail metrics~\cite{jeunen2024powerful}, engineers must identify performance gains~\cite{deng2016data}, assess statistical significance, attribute metric changes~\cite{xu2018sqr}, and balance potential risks. Analyzing a typical experiment and determining the next parameter configuration often requires several hours of work by an experienced engineer~\cite{fabijan2018effective}.
Meanwhile, historical strategy knowledge~\cite{namaki2020vamsa} is fragmented across heterogeneous sources~\cite{schlegel2023management}, including documents, source code, experiment configurations, and metric logs. These sources lack a unified abstraction~\cite{re2019overton} of strategy mechanisms, applicability conditions, and transferability boundaries, making historical knowledge difficult to systematically retrieve and reuse. Even when a similar historical strategy is identified, differences in user intent, ranking pipelines, feature availability, and business constraints~\cite{sheng2021one} require its parameters, thresholds, and traffic allocation to be adapted to the target scenario~\cite{zhang2022scenario, wang2023plate} and continuously refined through multiple rounds of online experimentation~\cite{wang2023plate}.

Recent advances in large language models~\cite{jiang2025hibench,ning2025survey} and retrieval-augmented generation~\cite{jiang2025qa,yuan2025mkg} provide new opportunities for understanding and reusing industrial strategy knowledge. Nevertheless, conventional RAG agents remain insufficient for complex strategy iteration~\cite{sarthi2024raptor,edge2024local}. Flat RAG systems primarily retrieve independent text chunks according to semantic similarity and cannot adequately model the hierarchical relationships among application scenarios, business pipelines, ranking stages, optimization objectives, parameter configurations, and experimental outcomes~\cite{yoran2024making,asai2024self}. This limitation restricts reliable knowledge transfer across scenarios and pipelines~\cite{huang2025retrieval,tao2025treerag}. Moreover, existing RAG agents~\cite{shinn2023reflexion} typically provide only one-shot strategy recommendations and lack a long-horizon optimization framework that connects strategy generation, online experimentation, metric analysis, and parameter updates. Consequently, as shown in Figure~\ref{fig:intro} (b), they cannot autonomously refine strategies from continuous feedback or progressively consolidate successful practices and failed experiments into an strategy knowledge base.

To address the aforementioned challenges, \textbf{\ourname} is proposed as an agentic framework for industrial recommendation strategy knowledge construction, strategy initialization, online optimization, and experience accumulation and updating. The framework transforms fragmented historical experiment records into structured and transferable strategy experiences. Given a new business request, it retrieves relevant historical knowledge under the target context and deployment constraints to initialize an executable strategy. After deployment, the system continuously analyzes core and guardrail metrics, compares successive experiment variants, and generates the strategy and parameter configuration for the next testing round. Validated outcomes are subsequently written back to the strategy experience tree, forming a closed loop of experience accumulation, strategy initialization, online tuning, and experience accumulation.

The framework consists of three core components, \textit{Historical Strategy Knowledge Organization}, \textit{Autonomous Target-Aware Strategy Generation}, and \textit{Experiment-Guided Strategy Self-Evolution}.
To overcome fragmented experience, as shown in Figure~\ref{fig:intro} (c), historical A/B records are decomposed into reusable strategy chunks containing motivations, mechanisms, parameters, applicable contexts, outcomes, and risks, and are organized into a hierarchical experience tree by domain, scenario, recommendation stage, and objective. To improve the weak knowledge transfer of flat RAG, the initialization module performs multi-path sparse and dense retrieval, hierarchy-aware candidate enhancement, and reranking, thereby selecting context-matched evidence and generating an executable strategy with feasible parameter settings and controlled guardrail risks.
After deployment, the iterative refinement and experience-updating module reduces manual and time-consuming tuning by organizing successive strategies, parameter updates, and online metrics into an experiment tree. Parent--child and sibling comparisons identify effective modifications and promising alternatives, enabling autonomous local parameter search or mechanism replacement when performance saturates, or guardrails are violated. Validated strategies are then written back to the experience tree, forming a closed loop of knowledge organization, target-aware initialization, autonomous A/B optimization, and experience self-updating.

Extensive offline and online evaluations demonstrate the effectiveness of \ourname{}. Across three industrial e-commerce scenarios, \ourname{} achieves the highest average overall score of \(8.126\), outperforming Claude-Sonnet-4.6 by \(21.2\%\). Compared with the strongest RAG baseline in each scenario, it improves the overall score by \(58.4\%\), \(66.2\%\), and \(54.4\%\), respectively. It also consistently surpasses GPT-5.5 in correctness and Claude-Sonnet-4.6 in novelty. In real-world deployment, \ourname{} further achieves a \textbf{4.829\%} improvement in Cart GMV while maintaining positive gains across all guardrail metrics.

The main contributions of this work are summarized as follows:

\begin{itemize}
\item \ourname{} is introduced as a closed-loop agent for historical strategy mining, cross-scenario transfer, A/B analysis, and parameter optimization in industrial recommender systems.

\item An experiment-tree-guided mechanism is developed for A/B tuning and experience self-evolution, enabling joint metric analysis, parameter optimization, and continuous knowledge updates.

\item Extensive offline and online experiments validate the effectiveness of \ourname, including a \textbf{4.829\%} improvement in GMV in real-world deployment.
\end{itemize}

\begin{figure*}
    \centering
    \includegraphics[width=\linewidth]{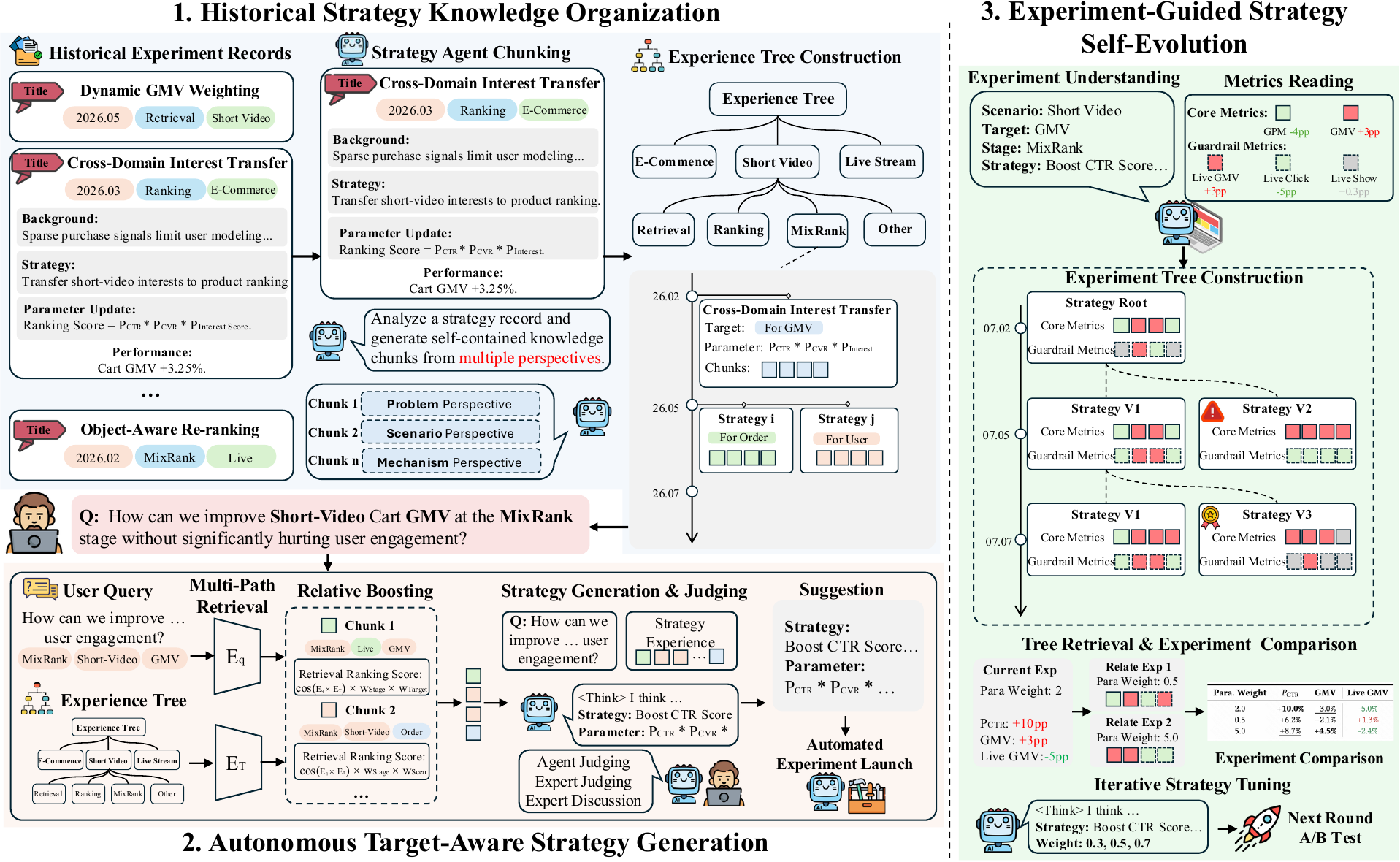}
    \caption{Overview of \ourname. The framework consists of three components: Historical Strategy Knowledge Organization, Autonomous Target-Aware Strategy Generation, and Experiment-Guided Strategy Self-Evolution.}
    \label{fig:framework}
\end{figure*}

\section{Related Work}

\textbf{Agents for Recommendation Optimization.}
Recent studies have explored agent-driven optimization loops for recommender systems, but differ in their optimization targets and experience reuse mechanisms.
AgentX~\cite{agentx2026selfiteration} automates end-to-end recommendation R\&D through multi-agent collaboration, including idea generation, code implementation, and online A/B testing, while using SGPO to refine agent prompts and execution frameworks.
AgentX-Model~\cite{yang2026agentxmodel} focuses on iterative model research through module reproduction, extension, and composition, primarily guided by offline AUC.
The self-evolving system in~\cite{wang2026selfevolvingrec} optimizes model architectures, optimizers, and reward functions through offline evaluation and online feedback, accumulating experience in an Experiment Journal.
AgenticRecTune~\cite{wu2026agenticrectune} similarly leverages live A/B feedback and a Skillhub for experience reuse, but primarily tunes numerical configurations such as fusion weights and diversity thresholds.
Unlike these approaches, \ourname{} keeps the foundation model, prompts, and tuning rules fixed, instead continuously constructing a hierarchical strategy experience tree from validated online A/B outcomes.
The tree organizes experiences by scenario, stage, and objective, while an experiment tree supports strategy variant comparison.
By retrieving transferable historical evidence to guide subsequent optimization and writing validated outcomes back into the tree, \ourname{} establishes an experience-driven self-evolution loop under core and guardrail metrics on real traffic.

\textbf{Sequential Strategy Optimization.}
Sequential optimization methods~\cite{sui2015safe}, including Bayesian optimization~\cite{NIPS2012_05311655}, contextual bandits~\cite{li2010contextual}, reinforcement learning~\cite{zheng2018drn}, and adaptive experimental allocation~\cite{simchi2023multi}, improve optimization efficiency by balancing exploration and exploitation through iterative feedback. However, these methods primarily focus on searching predefined parameter spaces, often overlooking underlying recommendation mechanisms, historical experimental knowledge, and complex relationships among business objectives. 
Moreover, they provide limited support for systematically extracting and reusing knowledge from historical experiments to guide strategy generation.
In industrial recommendation systems, costly online A/B testing and sparse feedback further constrain extensive exploration and repeated trial-and-error. In contrast, \ourname{} adopts a knowledge-driven approach that integrates historical strategy experiences, recommendation mechanisms, and business constraints to generate targeted and interpretable optimization strategies. These strategies are continuously validated and refined through online A/B feedback, enabling more effective and knowledge-guided strategy iteration.

\section{Methodology}
\label{sec:methodology}

\begin{figure*}
    \centering
    \includegraphics[width=\linewidth]{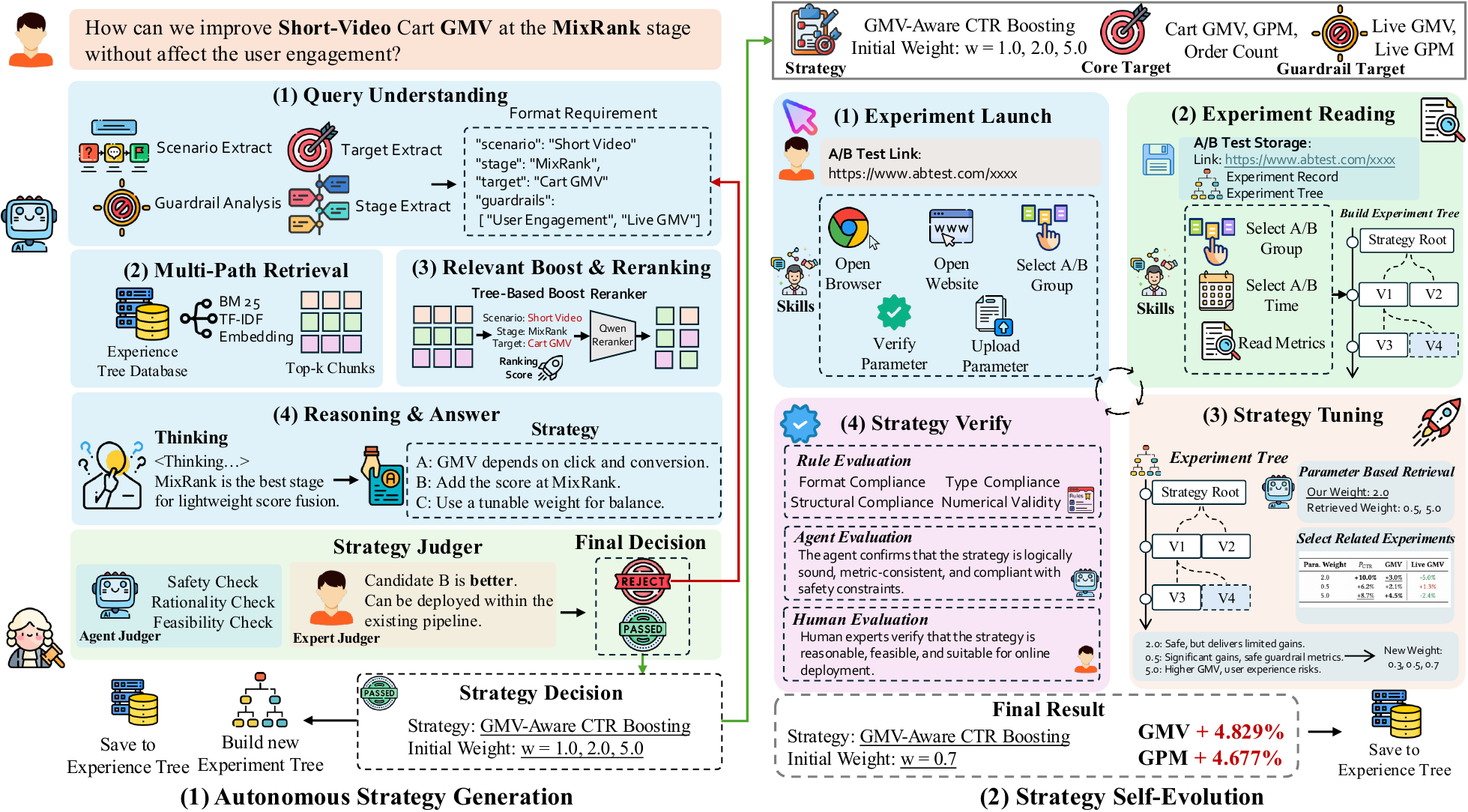}
   \caption{End-to-end workflow of \ourname{}, including autonomous strategy generation via target-aware multi-path retrieval and strategy self-evolution through online A/B experimentation, feedback-driven tuning, and continuous experience updates.}
    \label{fig:delpoy}
\end{figure*}

As illustrated in Figure~\ref{fig:framework}, \ourname{} consists of three components: \textit{Historical Strategy Knowledge Organization}, \textit{Autonomous Target-Aware Strategy Generation}, and \textit{Experiment-Guided Strategy Self-Evolution}. Historical experiment records are first transformed into reusable strategy experiences and organized into a hierarchical experience tree. Given a new optimization request, Tree-RAG retrieves relevant experiences according to the business context and generates an executable initial strategy. After deployment, successive A/B experiments are organized into an experiment tree, where strategy and parameter comparisons guide iterative optimization.

\subsection{Historical Strategy Knowledge Organization}

\label{sec:experience_tree}

\paragraph{Historical Experiment Record Structuring.}
Historical experiment reports often contain multiple strategy changes, parameter versions, and intermediate failures, making document-level retrieval prone to irrelevant information. We standardize each record into a unified schema covering business context, recommendation stage, optimization objectives, strategy modifications, performance metrics, and deployment conditions. This structured representation preserves essential experimental details while providing a consistent foundation for strategy extraction, cross-experiment comparison, and experience tree construction.

\paragraph{Agentic Multi-Perspective Strategy Extraction.}
A strategy agent decomposes each record into atomic, self-contained strategy chunks, preserving their motivations, mechanisms, parameter configurations, applicability conditions, outcomes, and risks. To support diverse retrieval queries, each chunk is represented from problem, scenario, and mechanism perspectives. Schema validation, metric consistency checks, and atomicity checks ensure the reliability and independence of extracted strategies.

\paragraph{Hierarchical Experience Tree Construction.}
The extracted chunks are organized into a hierarchical strategy experience tree $\mathcal{T}_{E}=(\mathcal{V}_{E},\mathcal{E}_{E})$, where nodes represent categories and strategies, and edges encode hierarchical relations. Each chunk $c$ is indexed by a semantic path: $\pi(c)=[d_c,s_c,l_c,o_c]$ corresponding to its domain, scenario, recommendation stage, and optimization objective. Each leaf stores strategy details, configurations, metric changes, deployment conditions, risks, and experiment references. Transferable strategies can be indexed under multiple paths, supporting precise retrieval and cross-branch knowledge transfer.

\paragraph{Multi-Path Strategy Retrieval.}
Given an optimization request $q$, \ourname{} identifies its business scenario, recommendation stage, optimization objective, guardrail requirements, and deployment constraints, and maps them to a semantic path in the strategy experience tree. Candidate strategy chunks are retrieved through complementary sparse and dense paths. Let $S_m(c,q)$ denote the relevance score between candidate chunk $c$ and request $q$ from retrieval path $m$. The initial retrieval score is
\begin{equation}
S_{\mathrm{ret}}(c,q)=
\sum_{m=1}^{M}
\alpha_m
\operatorname{Norm}\left(S_m(c,q)\right),
\end{equation}
where $M$ is the number of retrieval paths, $\alpha_m$ is the corresponding weight, and $\operatorname{Norm}(\cdot)$ aligns score ranges across methods. Sparse retrieval captures exact business terms, metric names, and parameter identifiers, while dense retrieval identifies strategies with similar mechanisms but different expressions. Their combination balances exact matching and semantic recall.

\paragraph{Context-Aware Relevance Boosting and Reranking.}
Since textual similarity alone may overlook business context, \ourname{} leverages the experience tree to boost candidates matching the target scenario, recommendation stage, and optimization objective. The tree-path relevance between candidate $c$ and request $q$ is defined as
\begin{equation}
B_{\mathrm{tree}}(c,q)=
\sum_l \beta_l
\exp\left(
-\gamma_l d_{\mathcal{T}}(v_c^l,v_q^l)
\right),
\end{equation}
where $d_{\mathcal{T}}$ denotes the shortest-path distance between nodes at hierarchy level $l$, while $\beta_l$ and $\gamma_l$ control the level importance and distance decay, respectively. This design prioritizes structurally relevant experiences while preserving cross-scenario transferability.
The tree-path relevance is combined with initial retrieval scores to select the top-$K$ candidates, which are further refined by Qwen-Reranker based on semantic relevance, applicability, and deployment constraints. The highest-ranked chunks are retained as evidence for strategy generation.

\paragraph{Evidence-grounded Strategy Generation and Agentic Judging.}
The strategy generator integrates optimization requests, business context, and retrieved experiences to produce executable strategies with concrete configurations and supporting evidence under deployment constraints. An LLM-based judging framework supports iterative refinement within \ourname{} and offline evaluation. Five specialized agents assess each strategy along five dimensions: (1) \textit{Correctness}, measuring technical soundness and scenario relevance; (2) \textit{Evidence Strength}, assessing grounding in historical evidence; (3) \textit{Delivery Completeness}, evaluating implementation details, evidence, and risk analysis; (4) \textit{Feasibility}, assessing compatibility with engineering constraints and guardrails; and (5) \textit{Novelty}, measuring insights beyond historical strategies. A separate holistic rubric produces an \textit{Overall} score independent of the five dimension scores. Scores and feedback guide iterative refinement and re-evaluation. Guardrails are enforced as hard constraints, and strategies that contradict historical evidence or violate critical deployment constraints are filtered out regardless of their Overall scores. The feasible strategy with the highest Overall score is selected for online A/B testing.

\subsection{Experiment-Guided Strategy Self-Evolution}
\label{sec:ab_tuning}

\paragraph{A/B Experiment Understanding and Outcome Extraction.}
After deployment, \ourname{} structures experiment descriptions, configurations, and metadata into unified records covering business context, optimization objectives, strategies, and guardrails. It extracts metric changes, experiment duration, traffic allocation, and statistical confidence. To mitigate noisy outcomes, each metric change is confidence-adjusted as
$
\widetilde{\Delta y}_j = \rho_j \Delta y_j,
$
where $\Delta y_j$ is the observed metric change and $\rho_j\in[0,1]$ denotes its confidence weight.

\paragraph{A/B Experiment Tree Construction.}
Successive experiments are organized into a tree $\mathcal{T}_A=(\mathcal{V}_A,\mathcal{E}_A)$, where nodes represent experiment versions and edges capture strategy or parameter modifications. Each node
$
v_t=(a_t,\boldsymbol{\theta}_t,\mathbf{y}_t,x_t)
$
records the tested strategy, configuration, metric outcomes, and context. The root represents the initial baseline, while child and sibling nodes capture successive modifications and alternative configurations, respectively.

\paragraph{Experiment Comparison and Utility Evaluation.}
For the current experiment, \ourname{} retrieves nodes with similar scenarios, objectives, and strategy mechanisms, and performs both parent--child and sibling comparisons. Parent--child comparison evaluates the effect of a specific modification, while sibling comparison identifies the most promising configuration. To jointly consider core gains and guardrail risks, the utility of an experiment node is defined as
\begin{equation}
U(v)=
\sum_{j\in\mathcal{M}_{\mathrm{core}}}
w_j\widetilde{\Delta y}_j
-
\lambda
\sum_{k\in\mathcal{M}_{\mathrm{guard}}}
w_k
\max\left(0,-\widetilde{\Delta y}_k\right),
\end{equation}
where $\mathcal{M}_{\mathrm{core}}$ and $\mathcal{M}_{\mathrm{guard}}$ denote the sets of core and guardrail metrics, respectively; $w_j$ and $w_k$ are their importance weights; and $\lambda$ controls the penalty for guardrail degradation. A higher $U(v)$ indicates a better balance between target improvement and guardrail safety.

\paragraph{Iterative Strategy Tuning and Experience Updating.}
Based on current outcomes, related experiment nodes, and historical experiences, \ourname{} recommends strategies and parameter configurations for subsequent A/B rounds. For promising branches, the agent performs local parameter search, while saturated or underperforming branches trigger the retrieval of alternative strategies from $\mathcal{T}_E$. After each round, both successful and unsuccessful experiments are converted into atomic strategy chunks through the same extraction and validation pipeline as historical records, preserving their configurations, contexts, and confidence-adjusted outcomes $\widetilde{\Delta y}_j$. Each chunk is assigned a semantic path $\pi(c)$, linked to its experiment node $v_t$ in $\mathcal{T}_A$, and incorporated into $\mathcal{T}_E$ for future reuse. This iterative process continues until improvements converge, the experiment budget is exhausted, or guardrail constraints prevent further exploration.
\section{Deployment}

Figure~\ref{fig:delpoy} illustrates the online deployment workflow of \ourname{}, consisting of two stages: \textit{Autonomous Strategy Generation} and \textit{Strategy Self-Evolution}. Built on GLM-5.1, \ourname{} identifies the target scenario, recommendation stage, optimization objectives, and deployment constraints from a strategy request. It then retrieves historical experiences via Tree-RAG and generates executable strategies with parameter configurations, supporting evidence, and risk assessments. Before deployment, rule-based validation and agent-human review ensure correctness, safety, and feasibility. Human experts only assess and intercept system-level risks, without modifying agent-generated strategies or intervening in benefit judgments.
After deployment, \ourname{} collects online A/B testing results and updates the experiment tree with new strategies and outcomes. By comparing parent and sibling nodes, it refines promising strategies or explores alternatives when performance saturates or guardrails are violated. Validated outcomes are incorporated into the strategy experience tree, enabling continuous knowledge accumulation.

\section{Experiments}

\subsection{Experimental Setup}
\begin{table}[t]
\centering
\caption{Dataset Statistics. Scenarios 1--3 are three presentation surfaces with different content carriers within the same app on an industrial platform.}
\label{tab:strategy_distribution}
\resizebox{\columnwidth}{!}{
\begin{tabular}{lrrrrr}
\toprule
\textbf{Scenario} &
\textbf{Retrieval} &
\textbf{Ranking} &
\textbf{MixRank} &
\textbf{Others} &
\textbf{Total} \\
\midrule
Scenario 1   & 27 & 31 & 5  & 29  & 92  \\
Scenario 2   & 23 & 22 & 0  & 70  & 115 \\
Scenario 3 & 11 & 23 & 5  & 64  & 103 \\
\midrule
\textbf{Total}         & \textbf{61} & \textbf{76} & \textbf{10} & \textbf{163} & \textbf{310} \\
\bottomrule
\end{tabular}
}
\end{table}

\begin{table*}[t]
\centering
\caption{
Main experimental results across three industrial recommendation
scenarios.
The best and second-best results are highlighted in
\textbf{bold} and \underline{underline}, respectively.
Corr., Evid., Deliv., Feas., and Novel. denote Correctness, Evidence Strength, Delivery Completeness, Feasibility, and Novelty, respectively, and Overall is the holistic score under the evaluation framework defined in the Methodology.
}
\label{tab:main_results}

\setlength{\tabcolsep}{3pt}
\renewcommand{\arraystretch}{1.15}

\resizebox{\textwidth}{!}{
\begin{tabular}{ll*{18}{c}@{\hspace{6pt}}c}
\toprule

\multirow{2}{*}{\textbf{Method}}
& \multirow{2}{*}{\textbf{Model}}
& \multicolumn{6}{c}{\textbf{Domain 1}}
& \multicolumn{6}{c}{\textbf{Domain 2}}
& \multicolumn{6}{c}{\textbf{Domain 3}}
& \textbf{Overall} \\

\cmidrule(lr){3-8}
\cmidrule(lr){9-14}
\cmidrule(lr){15-20}
\cmidrule(l){21-21}

& 
& Corr. & Evid. & Deliv. & Feas. & Novel. & Overall
& Corr. & Evid. & Deliv. & Feas. & Novel. & Overall
& Corr. & Evid. & Deliv. & Feas. & Novel. & Overall
& Avg. \\

\midrule

\multirow{5}{*}{LLM}
& GLM-5.1
& 6.009 & 4.487 & 4.930 & 5.165 & 4.948 & 5.214
& 6.065 & 4.478 & 5.043 & 5.337 & 4.946 & 5.286
& 6.010 & 4.583 & 5.039 & 5.282 & 4.961 & 5.280
& 5.257 \\

& DeepSeek-V4-Pro
& 5.826 & 4.530 & 5.974 & 5.600 & 5.461 & 5.515
& 5.848 & 4.565 & 5.913 & 5.609 & 5.304 & 5.502
& 5.583 & 4.534 & 5.903 & 5.612 & 5.340 & 5.418
& 5.479 \\

& Kimi-K2.6
& 6.470 & 4.826 & 5.661 & 5.922 & 5.591 & 5.782
& 6.435 & 4.620 & 5.576 & 5.859 & 5.457 & 5.687
& 6.233 & 4.660 & 5.573 & 5.816 & 5.369 & 5.617
& 5.699 \\

& GPT-5.5
& 6.835 & 5.348 & 5.887 & 6.026 & 5.670 & 6.070
& \underline{6.935} & \underline{5.283} & 5.783 & 6.033 & 5.630 & 6.063
& \underline{6.942} & 5.320 & 5.854 & 6.000 & 5.650 & 6.083
& 6.072 \\

& Claude-Sonnet-4.6
& \underline{6.965} & 5.330 & \underline{7.617} & \underline{7.165} & \underline{6.852} & \underline{6.797}
& 6.739 & 5.196 & \underline{7.652} & \underline{7.054} & \underline{6.685} & \underline{6.671}
& 6.660 & 5.301 & \underline{7.505} & \underline{7.000} & \underline{6.709} & \underline{6.630}
& \underline{6.704} \\

\midrule

\multirow{5}{*}{RAG Agent}
& Vanilla RAG
& 5.435 & 5.070 & 4.548 & 4.817 & 5.000 & 5.017
& 5.370 & 5.076 & 4.370 & 4.652 & 4.957 & 4.926
& 5.485 & 5.146 & 4.573 & 4.864 & 5.068 & 5.069
& 5.007 \\

& GAR
& 5.417 & 5.235 & 4.452 & 4.817 & 5.061 & 5.032
& 5.304 & 4.978 & 4.380 & 4.707 & 4.859 & 4.890
& 5.544 & 5.243 & 4.621 & 4.835 & 4.961 & 5.099
& 5.012 \\

& HyDE
& 5.522 & 5.322 & 4.591 & 4.870 & 5.174 & 5.130
& 5.293 & 5.022 & 4.370 & 4.630 & 4.870 & 4.879
& 5.495 & 5.175 & 4.495 & 4.864 & 4.981 & 5.053
& 5.030 \\

& CRAG
& 5.443 & \underline{5.374} & 4.435 & 4.783 & 5.191 & 5.070
& 5.359 & 5.022 & 4.337 & 4.685 & 4.946 & 4.911
& 5.612 & \underline{5.408} & 4.728 & 5.010 & 5.146 & 5.227
& 5.075 \\

& Graph RAG
& 5.452 & 5.200 & 4.504 & 4.887 & 5.078 & 5.062
& 5.293 & 5.011 & 4.326 & 4.696 & 4.967 & 4.891
& 5.417 & 5.097 & 4.515 & 4.796 & 4.971 & 5.004
& 4.992 \\

\midrule

\ourname{}
& \textbf{Ours}
& \textbf{7.765} & \textbf{7.817} & \textbf{8.809} & \textbf{8.243} & \textbf{8.235} & \textbf{8.127}
& \textbf{7.946} & \textbf{7.772} & \textbf{8.793} & \textbf{8.337} & \textbf{8.239} & \textbf{8.188}
& \textbf{7.660} & \textbf{7.777} & \textbf{8.835} & \textbf{8.165} & \textbf{8.165} & \textbf{8.070}
& \textbf{8.126} \\

\bottomrule
\end{tabular}
}
\end{table*}

\newcommand{\abci}[3]{%
    \mbox{%
        #1%
        {\scriptsize\textcolor{gray}{\,[#2,\,#3]}}%
    }%
}

\begin{table*}[t]
\centering
\caption{
Online A/B testing results and tuning efficiency comparison between expert-designed and agent-generated strategies. Confidence intervals are reported as subscripts following the corresponding metric improvements. Each round of experiments was conducted on 10\% of real production traffic, always against a baseline control group of the same traffic size.
}
\label{tab:online_ab_results}

\setlength{\tabcolsep}{3pt}
\renewcommand{\arraystretch}{1.20}

\resizebox{\textwidth}{!}{%
\begin{tabular}{
    l
    l
    c
    @{\hspace{6pt}}
    c
    @{\hspace{6pt}}
    cccc
    @{\hspace{6pt}}
    cccc
}
\toprule

\multirow{2}{*}{\textbf{Method}}
&
\multirow{2}{*}{\textbf{Strategy}}
&
\multirow{2}{*}{\textbf{Tuning Time}}
&
\multicolumn{1}{c}{\textbf{Core Metric}}
&
\multicolumn{4}{c}{\textbf{Important Metrics}}
&
\multicolumn{4}{c}{\textbf{Guardrail Metrics}}
\\

\cmidrule(lr){4-4}
\cmidrule(lr){5-8}
\cmidrule(lr){9-12}

&
&
&
\textbf{GMV}
&
\textbf{GPM}
&
\textbf{OPM}
&
\textbf{Clicks}
&
\textbf{Orders}
&
\textbf{Watch Time}
&
\textbf{Live GMV}
&
\textbf{Plat. GMV}
&
\textbf{Plat. Orders}
\\

\midrule

\rowcolor{mygray}
\textbf{Expert}
& Final
& \textbf{14 days}
& \abci{+2.679\%}{-0.22}{5.58}
& \abci{+1.736\%}{-1.26}{4.74}
& \abci{+1.492\%}{-0.15}{3.13}
& \cellcolor{secondred}\abci{\textbf{+0.872\%}}{0.35}{1.39}
& \cellcolor{secondred}\abci{\textbf{+2.250\%}}{0.65}{3.85}
& \cellcolor{negativegreen}\abci{-0.093\%}{-0.26}{0.07}
& \cellcolor{negativegreen}\abci{-0.636\%}{-1.99}{0.71}
& \cellcolor{positivered}\abci{+0.399\%}{-0.26}{1.06}
& \cellcolor{positivered}\abci{+0.026\%}{-0.31}{0.36}
\\

\midrule

\multirow{5}{*}{\textbf{Agent}}
& Strategy 1
& 1 day
& \cellcolor{lightgray}\abci{+1.123\%}{-3.38}{5.62}
& \cellcolor{lightgray}\abci{+0.545\%}{-3.94}{5.04}
& \cellcolor{lightgray}\abci{+1.001\%}{-0.39}{2.39}
& \cellcolor{lightgray}\abci{+0.557\%}{0.12}{1.00}
& \cellcolor{lightgray}\abci{+1.543\%}{0.17}{2.91}
& \cellcolor{negativegreen}\abci{-0.067\%}{-0.20}{0.07}
& \cellcolor{negativegreen}\abci{-1.245\%}{-2.60}{0.11}
& \cellcolor{negativegreen}\abci{-0.355\%}{-1.04}{0.33}
& \cellcolor{positivered}\abci{+0.048\%}{-0.21}{0.31}
\\

\addlinespace[2pt]

& Strategy 2
& 1 day
& \cellcolor{lightgray}\abci{+2.566\%}{0.20}{4.93}
& \cellcolor{lightgray}\abci{+1.784\%}{-0.63}{4.20}
& \cellcolor{lightgray}\abci{+0.700\%}{-0.91}{2.31}
& \cellcolor{bestred}\abci{\textbf{+1.235\%}}{0.69}{1.78}
& \cellcolor{lightgray}\abci{+1.578\%}{-0.01}{3.17}
& \cellcolor{negativegreen}\abci{-0.033\%}{-0.20}{0.13}
& \cellcolor{negativegreen}\abci{-1.136\%}{-2.51}{0.24}
& \cellcolor{negativegreen}\abci{-0.115\%}{-0.78}{0.55}
& \cellcolor{negativegreen}\abci{-0.086\%}{-0.41}{0.24}
\\

\addlinespace[2pt]

& Strategy 3
& 2 days
& \cellcolor{lightgray}\abci{+2.902\%}{0.38}{5.43}
& \cellcolor{secondred}\abci{\textbf{+2.773\%}}{0.15}{5.39}
& \cellcolor{bestred}\abci{\textbf{+2.458\%}}{0.85}{4.06}
& \cellcolor{lightgray}\abci{+0.460\%}{-0.09}{1.01}
& \cellcolor{bestred}\abci{\textbf{+2.674\%}}{1.10}{4.25}
& \cellcolor{positivered}\abci{+0.085\%}{-0.08}{0.25}
& \cellcolor{negativegreen}\abci{-0.108\%}{-1.57}{1.35}
& \cellcolor{positivered}\abci{+0.639\%}{-0.03}{1.31}
& \cellcolor{positivered}\abci{+0.176\%}{-0.14}{0.49}
\\

\addlinespace[2pt]

& Strategy 4
& 2 days
& \cellcolor{secondred}\abci{\textbf{+3.253\%}}{1.41}{5.10}
& \cellcolor{lightgray}\abci{+2.742\%}{0.84}{4.65}
& \cellcolor{secondred}\abci{\textbf{+1.633\%}}{0.47}{2.79}
& \cellcolor{lightgray}\abci{+0.792\%}{0.38}{1.21}
& \cellcolor{lightgray}\abci{+2.194\%}{1.04}{3.34}
& \cellcolor{positivered}\abci{+0.022\%}{-0.10}{0.15}
& \cellcolor{positivered}\abci{+0.294\%}{-0.78}{1.37}
& \cellcolor{positivered}\abci{+0.314\%}{-0.19}{0.82}
& \cellcolor{positivered}\abci{+0.033\%}{-0.22}{0.29}
\\

\addlinespace[2pt]

& Strategy 5
& 2 days
& \cellcolor{bestred}\abci{\textbf{+4.829\%}}{2.38}{7.27}
& \cellcolor{bestred}\abci{\textbf{+4.677\%}}{2.26}{7.10}
& \cellcolor{lightgray}\abci{+0.841\%}{-0.61}{2.29}
& \cellcolor{lightgray}\abci{+0.370\%}{-0.18}{0.92}
& \cellcolor{lightgray}\abci{+1.053\%}{-0.43}{2.54}
& \cellcolor{positivered}\abci{+0.078\%}{-0.08}{0.23}
& \cellcolor{positivered}\abci{+0.812\%}{-0.43}{2.05}
& \cellcolor{positivered}\abci{+0.188\%}{-0.38}{0.76}
& \cellcolor{positivered}\abci{+0.288\%}{-0.01}{0.59}
\\

\midrule

\multicolumn{2}{l}{
    \textbf{$\Delta$ (Strategy 5 $-$ Expert)}
}
& \textbf{$1.75\times$}
& \textbf{+2.150 pp}
& \textbf{+2.941 pp}
& -0.651 pp
& -0.502 pp
& -1.197 pp
& \textbf{+0.171 pp}
& \textbf{+1.448 pp}
& -0.211 pp
& \textbf{+0.262 pp}
\\

\bottomrule
\end{tabular}%
}

\vspace{3pt}
\begin{minipage}{\textwidth}
\footnotesize
\textit{Note:}
Values denote relative A/B metric changes, with CIs in subscript brackets. Dark/light red indicate the best/second-best core and important results; red/green denote positive/negative guardrail changes. The final row reports pp differences between Strategy 5 and the expert strategy. Tuning efficiency compares 8 days for five agent strategies with 14 days for expert tuning, yielding a $1.75\times$ speedup.
\end{minipage}

\end{table*}
\textbf{Dataset.}
We construct an industrial strategy generation benchmark from 310 historical strategies across three scenarios of a large-scale e-commerce recommendation platform. The three scenarios correspond to three positions of the same APP with different content carriers, user intent, content form, and optimization context.
As summarized in Table~\ref{tab:strategy_distribution}, the benchmark covers diverse deployment stages, including retrieval, ranking, blending, and other recommendation components. Its broad coverage across scenarios and pipeline stages enables a comprehensive evaluation of strategy understanding and generation in heterogeneous industrial settings.

\textbf{Baselines.}
\ourname{} is compared with two groups of baselines. The first group includes general-purpose LLMs: GLM-5.1, DeepSeek-V4-Pro, Kimi-K2.6, GPT-5.5, and Claude-Sonnet-4.6, which directly generate strategies from the given scenario and task description. The second group consists of representative retrieval-augmented generation methods, including Vanilla RAG, GAR~\cite{mao2021generation}, HyDE~\cite{gao2023precise}, CRAG~\cite{yan2024corrective}, and Graph RAG~\cite{edge2024local}, which retrieve relevant historical experience before generating the final strategy. For a controlled comparison, all RAG baselines and \ourname{} use GLM-5.1 as the shared foundation model, ensuring that performance differences primarily stem from their knowledge organization, retrieval, and strategy-generation mechanisms.

\textbf{Evaluation Protocol and Validity of LLM-Based Evaluation}

For each case, a strategy is generated based on the target scenario, objective, and context. To prevent information leakage and look-ahead bias, we enforce strict \textbf{temporal isolation}, allowing access only to historical experiences available before the target experiment. All offline scores are produced by an independent external evaluator, DeepSeek-V4.1-Flash, which uniformly applies the five evaluation dimensions defined in the Methodology---Correctness, Evidence Strength, Delivery Completeness, Feasibility, and Novelty---together with a holistic Overall score. The internal judging agents of \ourname{} are used exclusively for pre-deployment strategy refinement, and their scores are excluded from the reported results.

To assess the reliability of the evaluation framework, we examine the association between offline LLM scores and online A/B testing outcomes. As shown in Table~\ref{tab:llm_online_correlation}, all evaluation dimensions exhibit positive correlations with online performance and help distinguish strategies with positive and negative impacts. Specifically, Correctness achieves the highest Pearson correlation of 0.742 and an AUC of 0.882, while the Overall score attains a Pearson correlation of 0.668 and an AUC of 0.870. These results suggest that the evaluation framework captures meaningful signals associated with online outcomes and provides a useful offline proxy for assessing strategy quality.

\begin{table}[t]
    \centering
    \caption{Correlation between LLM evaluation scores and online A/B outcomes. Strategies are grouped by core-metric improvement sign; Positive and Negative denote the corresponding mean LLM scores.}
    \label{tab:llm_online_correlation}
    \setlength{\tabcolsep}{6pt}
    \renewcommand{\arraystretch}{1.12}
    \resizebox{\columnwidth}{!}{
    \begin{tabular}{lcccc}
        \toprule
        \multirow{2}{*}{\textbf{Evaluation Dimension}}
        & \multicolumn{2}{c}{\textbf{LLM score avg.}}
        & \multirow{2}{*}{\textbf{Pearson $r$}}
        & \multirow{2}{*}{\textbf{AUC}} \\
        \cmidrule(lr){2-3}
        & \textbf{Positive} & \textbf{Negative} & & \\
        \midrule
        Correctness
        & 7.33 & 3.33
        & 0.742 & 0.882 \\

        Evidence Strength
        & 7.15 & 5.35
        & 0.609 & 0.843 \\

        Delivery Completeness
        & 4.07 & 2.60
        & 0.527 & 0.819 \\

        Feasibility
        & 7.47 & 6.15
        & 0.488 & 0.766 \\

        \midrule
        \textbf{Overall}
        & 6.04 & 4.19
        & \textbf{0.668} & \textbf{0.870} \\
        \bottomrule
    \end{tabular}
    }
\end{table}

\subsection{Offline Results}

Table~\ref{tab:main_results} presents the offline results across three industrial recommendation scenarios. \ourname{} consistently achieves the best performance across all evaluation dimensions, demonstrating its effectiveness across heterogeneous recommendation tasks. Among standalone LLMs, Claude-Sonnet-4.6 achieves the highest average overall score of 6.704, while GPT-5.5 shows competitive correctness. In contrast, RAG-based agents generally underperform standalone LLMs, with CRAG achieving the highest average score of only 5.075, suggesting that simply retrieving historical knowledge is insufficient for effective strategy generation.
Overall, \ourname{} achieves an average score of 8.126, outperforming the strongest standalone LLM and RAG baseline by 21.2\% and 60.1\%, respectively. Across three scenarios, \ourname{} improves the overall score over the strongest standalone LLM by 19.6\%, 22.7\%, and 21.7\%, respectively. Furthermore, it achieves consistent improvements in both correctness and novelty, demonstrating its ability to generate technically reliable and innovative strategies. These gains highlight the importance of structured industrial knowledge and transferability-aware retrieval for effective strategy optimization.

\subsection{Online A/B Results and Iterative Refinement}
Table~\ref{tab:online_ab_results} presents online A/B testing results across five iterations, demonstrating how the agent progressively refines strategies to improve business objectives while balancing platform-level guardrails.
Initially, Strategy~2 improves GMV by 2.566\% and achieves the highest click gain of 1.235\%, but negatively affects all guardrail metrics. Strategy~3 further improves OPM and orders by 2.458\% and 2.674\%, respectively, while recovering most guardrail metrics. Through subsequent refinement, Strategy~5 achieves the largest GMV and GPM improvements of 4.829\% and 4.677\%, respectively, with positive changes across all guardrail metrics.
Compared with the expert-designed strategy, Strategy~5 achieves additional GMV and GPM gains of 2.150 and 2.941 percentage points, while improving Watch Time, Live GMV, and Platform Orders by 0.171, 1.448, and 0.262 percentage points, respectively. Despite minor declines in some secondary metrics, it delivers stronger core business gains with better guardrail balance.  Moreover, \ourname{} completes five tuning iterations in 8 days, compared with 14 days for expert tuning, achieving a 1.75$\times$ speedup. 
The gap is also evident in tuning cost: human experts spent 10 tuning iterations over 14 days, repeatedly adjusting parameters until the gains converged, yet reached only a suboptimal outcome (GMV +2.679\%), whereas the agent achieved a stronger result (GMV +4.829\%) with only 5 iterations in 8 days.

\subsection{Ablation Study}

\begin{table}[!t]
\centering
\caption{Ablation study on key designs of \ourname{}. Dimension abbreviations follow the evaluation framework defined in the Methodology and Table~\ref{tab:main_results}.}
\label{tab:ablation_study}
\setlength{\tabcolsep}{3pt}
\renewcommand{\arraystretch}{1.1}
\resizebox{\columnwidth}{!}{%
\begin{tabular}{lccccccc}
\toprule
\textbf{Setting} &
\textbf{Corr.} &
\textbf{Evid.} &
\textbf{Deliv.} &
\textbf{Feas.} &
\textbf{Novel.} &
\textbf{Overall} &
\textbf{Decline} \\
\midrule

\rowcolor{rowgray}
\textbf{\ourname{}}
& \textbf{7.765}
& \textbf{7.817}
& \textbf{8.809}
& \textbf{8.243}
& \textbf{8.235}
& \textbf{8.127}
& -- \\
\midrule

\multicolumn{8}{l}{\textbf{A. Knowledge Base}} \\
\quad w/o Knowledge Base
& 7.197 & 5.746 & 5.352 & 6.634
& 6.824 & 6.388
& \textcolor{red}{1.739$\downarrow$} \\
\midrule

\multicolumn{8}{l}{\textbf{B. Retrieval Architecture}} \\
\quad w/o Strategy-Tree RAG
& 7.317 & 6.415 & 6.204 & 7.218
& 7.127 & 6.875
& \textcolor{red}{1.252$\downarrow$} \\

\quad w/o Strategy-Tree RAG (v2)
& 7.225 & 6.338 & 6.169 & 7.176
& 7.176 & 6.822
& \textcolor{red}{1.305$\downarrow$} \\

\quad w/o Multi-Path Retrieval
& 7.211 & 6.296 & 5.915 & 6.993
& 7.042 & 6.708
& \textcolor{red}{1.419$\downarrow$} \\
\midrule

\multicolumn{8}{l}{\textbf{C. Retrieval Enhancement}} \\
\quad w/o Objective-Aware Boosting
& 7.331 & 6.387 & 5.915 & 7.113
& 7.092 & 6.792
& \textcolor{red}{1.335$\downarrow$} \\

\quad w/o Scenario-Aware Boosting
& 7.246 & 6.394 & 5.880 & 7.028
& 7.049 & 6.739
& \textcolor{red}{1.388$\downarrow$} \\

\quad w/o Re-ranking
& 7.218 & 6.444 & 5.817 & 7.028
& 7.014 & 6.725
& \textcolor{red}{1.402$\downarrow$} \\

\bottomrule
\end{tabular}%
}
\end{table}

Table~\ref{tab:ablation_study} presents the ablation results in the short-video scenario. Removing any component consistently degrades performance, demonstrating the complementary contributions of the proposed designs. In particular, removing the knowledge base leads to the largest overall score decline of 1.739 (21.4\%), highlighting the importance of historical industrial experience. Replacing Strategy-Tree RAG with alternative retrieval architectures reduces the overall score by 1.252--1.305, while removing multi-path retrieval causes a larger decline of 1.419. These results demonstrate the benefits of hierarchical knowledge organization and diverse evidence retrieval.
Furthermore, removing objective-aware boosting, scenario-aware boosting, and re-ranking decreases the overall score by 1.335, 1.388, and 1.402, respectively. This indicates that effective knowledge transfer requires not only retrieving relevant experiences but also aligning them with target objectives and scenarios. Overall, the complete \ourname{} achieves the best performance across all evaluation dimensions, validating the effectiveness of its integrated knowledge retrieval and refinement mechanisms.

\subsection{Evaluation Consistency Across LLM Judges}
\label{sec:evaluation_consistency}

\begin{table}[b]
    \centering
    \caption{Evaluation consistency across different LLM judges.}
    \label{tab:judge_consistency}
    \resizebox{0.85\columnwidth}{!}{
    \begin{tabular}{lccc}
        \toprule
        \textbf{Evaluator} & \textbf{Claude} & \textbf{GPT} & \textbf{\ourname{}} \\
        \midrule
        DeepSeek-V4.1-Flash & 6.797 & 6.070 & \textbf{8.127} \\
        Qwen3-8-Flash       & 6.924 & 6.213 & \textbf{8.394} \\
        GLM-5-3-Flash       & 7.671 & 6.715 & \textbf{8.819} \\
        MiniMax-M3.1-Flash  & 6.322 & 6.652 & \textbf{7.576} \\
        \midrule
        \rowcolor{rowgray} \textbf{Human Evaluation}    & 7.608 & 7.148 & \textbf{8.235} \\
        \bottomrule
    \end{tabular}
    }
\end{table}

To further validate the reliability and robustness of our evaluation framework, we employ four independent LLM judges with diverse model architectures and conduct human evaluation as an additional reference. As shown in Table~\ref{tab:judge_consistency}, although different judges exhibit variations in their absolute scoring scales and relative preferences for baseline methods, all four LLM judges consistently rank \ourname{} above Claude and GPT. Specifically, \ourname{} achieves scores ranging from 7.576 to 8.819, outperforming the strongest baseline under each judge by 0.924--1.470 points. Notably, even when the relative rankings of Claude and GPT vary across evaluators, \ourname{} consistently maintains its leading position, suggesting that its superiority is not dependent on any particular LLM judge or scoring preference. Furthermore, human evaluation corroborates these findings, with \ourname{} achieving the highest score of 8.235, compared with 7.608 and 7.148 for Claude and GPT, respectively. The agreement between LLM and human evaluations further confirms the robustness and reliability of our findings.

\subsection{Effect of Accumulated Experience over Time}
\label{sec:self_evolution}

To investigate how performance evolves with the accumulation of historical experiences, we performed a chronological evaluation in Domain 3, dividing tasks into three phases. As shown in Table~\ref{tab:self_evolution}, the average number of accumulated experiences increases from 70.1 to 576.3, while the performance gain over the GLM-5.1 baseline rises from 2.927 to 3.161. Meanwhile, \ourname achieves its highest Overall score of 8.419 in Phase 3, suggesting that a richer experience memory can benefit subsequent strategy generation. However, as task sets vary across phases, the observed improvements reflect both the accumulation of experience and the specific differences of the task, rather than the isolated effect of the experience-updating mechanism.

\definecolor{gain1}{RGB}{252,240,240}
\definecolor{gain2}{RGB}{248,219,219}
\definecolor{gain3}{RGB}{242,190,190}

\begin{table}[t]
    \centering
    \caption{Performance improvement across three phases of self-evolution.}
    \label{tab:self_evolution}
    \resizebox{0.9\columnwidth}{!}{
    \begin{tabular}{lcccc}
        \toprule
        \textbf{Phase}
        & \textbf{Avg. Exp.}
        & \textbf{A/B Agent}
        & \textbf{Baseline}
        & \textbf{Gain} \\
        \midrule
        Phase 1 & 70.1  & 8.333 & 5.407
        & \cellcolor{gain1}+2.927 \\
        Phase 2 & 313.3 & 8.258 & 5.316
        & \cellcolor{gain2}+2.942 \\
        Phase 3 & 576.3 & \textbf{8.419} & 5.258
        & \cellcolor{gain3}\textbf{+3.161} \\
        \bottomrule
    \end{tabular}
    }
\end{table}

\subsection{Cross-Scenario Knowledge Transfer}
\label{sec:cross_scenario}

\begin{table}[b]
    \centering
    \caption{Performance comparison on tasks involving cross-scenario retrieval.}
    \label{tab:cross_scenario_retrieval}
    \setlength{\tabcolsep}{5pt}
    \renewcommand{\arraystretch}{1.15}
    \resizebox{\columnwidth}{!}{
    \begin{tabular}{lcccc}
        \toprule
        \textbf{Target Scenario}
        & \textbf{\shortstack[c]{Cross-scenario\\HitRate}}
        & \textbf{\ourname{}}
        & \textbf{GLM-5.1}
        & \textbf{$\Delta$ Score} \\
        \midrule
        Scenairo 1
        & 7.8\% & \textbf{8.244} & 5.578 & +2.667 \\
        Scenairo 2
        & 51.1\% & \textbf{8.209} & 5.226 & +2.983 \\
        Scenairo 3
        & 47.6\% & \textbf{8.027} & 5.294 & +2.733 \\
        \bottomrule
    \end{tabular}
    }
\end{table}

To evaluate the cross-scenario knowledge transfer capability of \ourname{}, we analyze tasks that retrieve historical experiences from other recommendation scenarios. As shown in Table~\ref{tab:cross_scenario_retrieval}, Cross-scenario HitRate (the proportion of strategy proposals in the target scenario that successfully retrieve relevant strategies from other scenarios) reaches 7.8\%, 51.1\%, and 47.6\% in three scenairos, respectively. On these tasks, \ourname{} consistently outperforms GLM-5.1 by 2.667, 2.983, and 2.733 points, respectively. These results demonstrate the effectiveness of \ourname{} on tasks involving cross-scenario retrieval, highlighting its ability to leverage transferable industrial knowledge across heterogeneous recommendation scenarios.

\subsection{Stage Analysis}

\begin{table}[t]
\centering
\caption{
Performance comparison across different recommendation stages.
The best and second-best results are highlighted in
\textbf{bold} and \underline{underline}, respectively.
}
\label{tab:stage}

\setlength{\tabcolsep}{5pt}
\renewcommand{\arraystretch}{1.15}

\resizebox{\columnwidth}{!}{
\begin{tabular}{llcccc}
\toprule

\textbf{Category}
& \textbf{Model}
& \textbf{Retrieval}
& \textbf{Ranking}
& \textbf{Blending}
& \textbf{Others} \\

\midrule

\multirow{5}{*}{LLM}
& GLM-5.1
& 5.220 & 5.288 & 5.295 & 5.212 \\

& DeepSeek-V4-Pro
& 5.612 & 5.372 & 5.558 & 5.324 \\

& Kimi-K2.6
& 5.668 & 5.722 & 5.842 & 5.505 \\

& GPT-5.5
& 6.077 & 6.094 & 6.200 & 5.815 \\

& Claude-Sonnet-4.6
& \underline{6.783}
& \underline{6.645}
& \underline{6.822}
& \underline{6.498} \\

\midrule

\multirow{5}{*}{RAG Agent}
& Vanilla RAG
& 4.926 & 5.083 & 5.064 & 4.924 \\

& GAR
& 4.931 & 5.065 & 5.100 & 4.946 \\

& HyDE
& 4.982 & 5.038 & 5.056 & 5.093 \\

& CRAG
& 5.051 & 5.100 & 5.115 & 5.010 \\

& Graph RAG
& 4.978 & 5.026 & 5.000 & 4.924 \\

\midrule

\rowcolor{rowgray}

A/B Agent
& \textbf{Ours}
& \textbf{8.140}
& \textbf{8.065}
& \textbf{8.214}
& \textbf{8.129} \\

\bottomrule
\end{tabular}
}
\end{table}

Table~\ref{tab:stage} compares different methods across four recommendation stages. \ourname{} consistently achieves the best performance in retrieval, ranking, blending, and other stages, outperforming the strongest baseline, Claude-Sonnet-4.6, by 20.0\%, 21.4\%, 20.4\%, and 25.1\%, respectively. While Claude-Sonnet-4.6 leads among standalone LLMs, RAG-based methods generally perform worse, suggesting that simple retrieval of historical experiences is insufficient for effective strategy optimization. These results demonstrate the effectiveness of \ourname{} across heterogeneous recommendation stages, highlighting the benefits of stage-aware knowledge organization and transferability-aware retrieval.

\section{Conclusion}

We propose \ourname{}, a self-evolving agentic framework for industrial recommendation strategy optimization. By organizing historical experiences into a hierarchical knowledge tree, retrieving transferable strategies through multi-path Tree-RAG, and iteratively refining strategies using online A/B feedback, \ourname{} enables knowledge-driven strategy optimization. Extensive offline experiments demonstrate consistent improvements over LLM and RAG baselines across diverse scenarios and recommendation stages. Real-world deployment further achieves a 4.829\% GMV improvement alongside gains in key business metrics. Future work will focus on improving requirement alignment, metric attribution, and automated feasibility verification to enable more reliable strategy refinement.

\bibliographystyle{ACM-Reference-Format}
\balance
\bibliography{citation}

\newpage
\appendix

\begin{figure}[h]
\centering
\includegraphics[width=\linewidth]{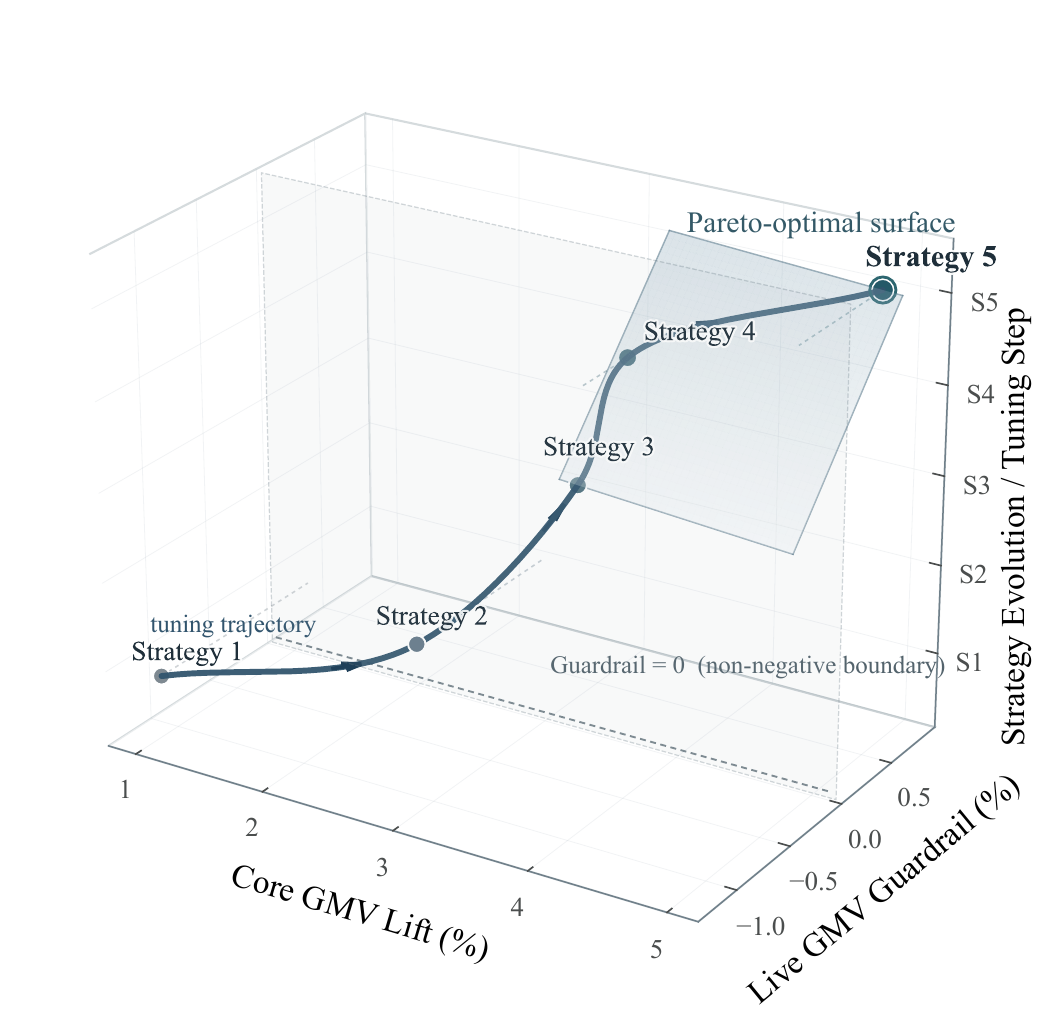}
\caption{Pareto-oriented tuning trajectory of \ourname{} across core gains and guardrail safety.}
\label{fig:pareto_tuning}
\end{figure}

\section{Balancing Core Gains and Guardrail Safety}

Figure~\ref{fig:pareto_tuning} illustrates how \ourname{} progressively balances core-metric gains and guardrail safety through online A/B feedback. Strategy~1 achieves a GMV improvement of \(+1.123\%\), but the average guardrail change is \(-0.405\%\). Strategy~2 raises GMV to \(+2.566\%\), while the average guardrail change remains negative at \(-0.343\%\).
Strategy~3 further increases GMV to \(+2.902\%\), with the average guardrail change turning positive at \(+0.198\%\). Strategy~4 improves GMV to \(+3.253\%\) and maintains positive changes across all four guardrail metrics, with an average improvement of \(+0.166\%\).
The final Strategy~5 achieves the largest GMV and GPM gains of \(+4.829\%\) and \(+4.677\%\), respectively, while improving all four guardrail metrics by an average of \(+0.342\%\). Compared with the expert strategy, it improves GMV and GPM by \(2.150\) and \(2.941\) percentage points (pp), respectively, alongside gains in Watch Time, Live GMV, and Platform Orders. Platform GMV decreases by \(0.211\) pp, indicating that trade-offs remain across individual metrics.
Overall, these results demonstrate that \ourname{} leverages online feedback to progressively improve business performance while maintaining positive guardrail outcomes, achieving a more favorable balance between business effectiveness and platform safety.

\section{Online A/B Test}
\begin{table}[t]
\centering
\caption{Online A/B testing results of \ourname.}
\label{tab:online_ab_summary}
\resizebox{\columnwidth}{!}{%
\begin{tabular}{lcccccc}
\toprule
\textbf{Method} &
\textbf{GMV} &
\textbf{GPM} &
\textbf{OPM} &
\textbf{CVR} &
\textbf{Clicks} &
\textbf{Orders} \\
\midrule
\textbf{\ourname}
& \textbf{+4.829\%}
& \textbf{+4.677\%}
& \textbf{+0.841\%}
& \textbf{+0.578\%}
& \textbf{+0.370\%}
& \textbf{+1.053\%} \\
\bottomrule
\end{tabular}%
}
\end{table}

\ourname{} has been deployed in the e-commerce short-video scenario on a large-scale recommendation platform. We conducted an online A/B test in a real-world production environment to evaluate its practical effectiveness. As shown in Table~\ref{tab:online_ab_summary}, the strategy generated by \ourname{} improves GMV by 4.829\%, together with gains of 4.677\% in GPM, 0.841\% in OPM, and 0.578\% in CVR. It also increases clicks and orders by 0.370\% and 1.053\%, respectively, demonstrating that \ourname{} can translate historical strategy knowledge into measurable online business improvements.

\section{Error Analysis}
\label{sec:error_analysis}

\begin{figure}[h]
    \centering
    \includegraphics[width=\linewidth]{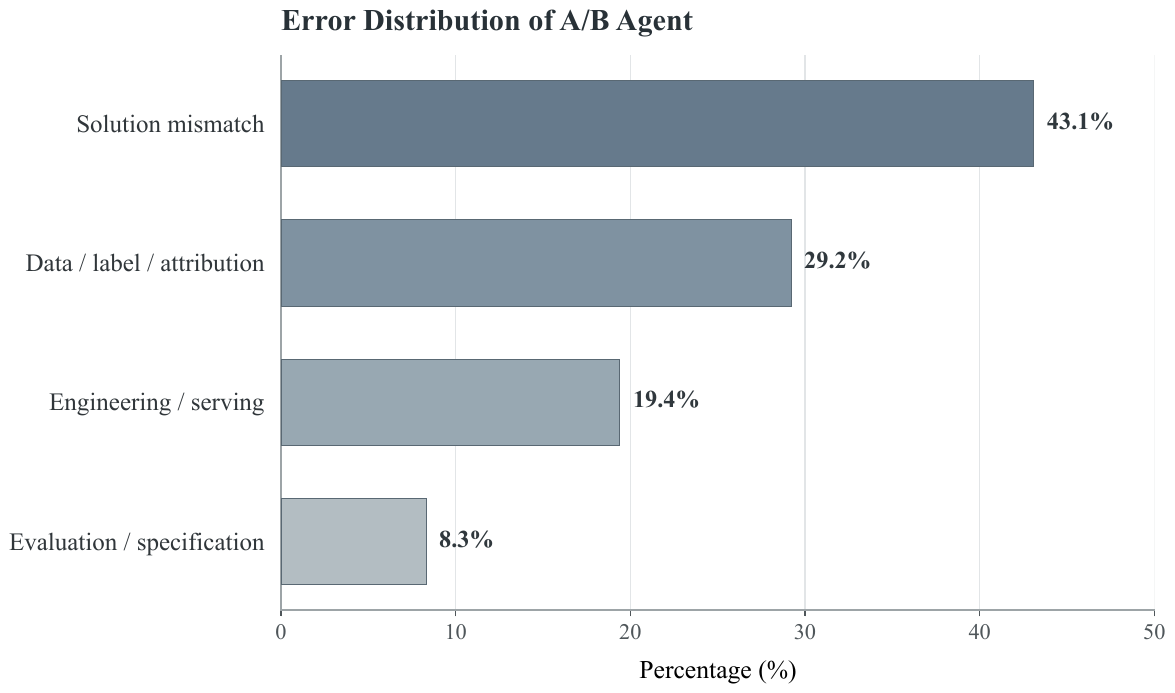}
    \caption{Distribution of error categories for \ourname.}
    \label{fig:error_analysis}
\end{figure}

Figure~\ref{fig:error_analysis} presents the distribution of failure cases produced by \ourname. The dominant error category is \emph{solution mismatch}, accounting for 43.1\% of all failures. These errors arise when a generated strategy is generally reasonable but does not fully align with the target scenario, optimization objective, pipeline stage, or deployment constraints. This result indicates that fine-grained strategy alignment remains the primary challenge for \ourname.

Data-, label-, and attribution-related errors constitute the second-largest category at 29.2\%. Such errors typically involve inaccurate assumptions about feature definitions, training labels, metric attribution, or the availability of historical data. Engineering and serving issues account for 19.4\%, mainly reflecting incompatibilities with existing pipelines, unavailable features, or excessive deployment overhead. Evaluation and specification errors represent only 8.3\% of the failures, suggesting that the task specification and evaluation process are comparatively reliable.

Overall, the remaining errors are primarily associated with scenario-specific understanding and practical constraint alignment rather than general strategy generation. Further improvements should therefore focus on objective-aware knowledge retrieval, explicit verification of data dependencies, and constraint-aware evaluation before online experimentation.

\end{document}